\documentclass[conference,onecolumn, 11pt]{IEEEtran}
\IEEEoverridecommandlockouts
\usepackage{cite}
\usepackage{amsmath,amssymb,amsfonts}
\usepackage{amsthm}
\usepackage{algorithm}
\usepackage{algorithmic}
\usepackage{graphicx}
\usepackage{textcomp}
\usepackage{xcolor}
\usepackage{cuted}
\usepackage{subcaption}
\usepackage{caption}
\usepackage{stfloats}

\usepackage{float}
\def\BibTeX{{\rm B\kern-.05em{\sc i\kern-.025em b}\kern-.08em
    T\kern-.1667em\lower.7ex\hbox{E}\kern-.125emX}}

\newcommand{\bbeta}{\boldsymbol\beta}

\newcommand{\btheta}{\boldsymbol\theta}

\newcommand{\bX}{\boldsymbol X}

\newcommand{\bx}{\boldsymbol x}
\newcommand{\by}{\boldsymbol y}
\newcommand{\bb}{\boldsymbol b}
\newcommand{\bk}{\boldsymbol k}
\usepackage{url}

\newtheorem{lemma}{Lemma}
\newtheorem{theorem}{Theorem}
\newtheorem{assumption}{Assumption}
\newtheorem{definition}{Definition}
\newtheorem{corollary}{Corollary}

\begin{document}

\title{
Identifying  Causal Direction via Dense \\Functional Classes}

\author{\IEEEauthorblockN{Kate\v rina Hlav\'a\v ckov\'a-Schindler}
\IEEEauthorblockA{ \textit{Research Group Data Mining and Machine Learning} \\
\textit{Faculty of Computer 
Science}\\
\textit{Data Science @ Uni Vienna} \\
\textit{University of Vienna,
Vienna, Austria} \\
katerina.schindlerova@univie.ac.at}
\and
\IEEEauthorblockN{Suzana Marsela}
\IEEEauthorblockA{\textit{Research Group Data Mining and Machine Learning} \\
\textit{Faculty of Computer 
Science, University of Vienna}\\
Vienna, Austria \\ suzana.marsela@gmail.com}
}

\maketitle

\begin{abstract}

We address the problem of determining the causal direction between two univariate, continuous-valued variables, 
$X$ and 
$Y$, under the assumption of no hidden confounders. In general, it is not possible to make definitive statements about causality without some assumptions on the underlying model. To distinguish between cause and effect, we propose a bivariate causal score based on the Minimum Description Length (MDL) principle, using functions that possess the density property on a compact real interval. We prove the identifiability of these causal scores under specific conditions.
These conditions can be easily tested. Gaussianity of the noise in the causal model equations is not assumed, only that the noise is low.
The well-studied class of cubic splines possesses the density property on a compact real interval. We propose  LCUBE as an instantiation of the MDL-based causal score utilizing cubic regression splines. LCUBE is an identifiable method that is also interpretable, simple, and  very fast. It has only one hyperparameter.
Empirical evaluations compared to state-of-the-art  methods demonstrate that LCUBE achieves superior precision in terms of AUDRC  on the real-world Tübingen cause-effect pairs dataset. It also shows superior average precision  across common 10 benchmark datasets  and  achieves above  average precision on 13 datasets.
\end{abstract}


\section{Introduction}\label{background}

We deal with a problem to decide about two univariate continuous-valued variables $X$ and $Y$ without hidden confounders, which variable is causing another one.
In
general, it is  not possible to make definite statements about causality
without some assumptions on the underlying model \cite{pearl2009causality}. 
Recently, Blöbaum et al. in \cite{bloebaum2018} formulated a set of assumptions under
which the true causal direction is identifiable via regression. They
showed that it is possible to identify cause from effect  by
selecting the direction with smaller residual error. The idea is to fit
both $Y = f (X ) + N_X$ and $X = g(X ) + N_Y$ by minimizing the respective
residual errors $N_X$ and $N_Y$, and then to compare the sum of
squared errors. 
They showed that, among models with complexity matching that of the true model, the model in the correct causal direction yields a lower error.
However, since the complexity of the true model is not know,  RECI  may compare residuals of arbitrarily underfit or overfit models.
Paper \cite{regularizedregression} builds on the results of \cite{bloebaum2018}, and formulates an additional condition  under which cause and effect are identifiable via  so-called regularized regression. 
The  so-called Identifiable Regression-based Scoring Function (= IRSF) is proposed and a general  framework  for the identifiability of these functions is presented, including an   instantiation by method SLOPPY,  using  splines and AIC and BIC as scoring function. 
Method Slope \cite{marx2017telling} also 
uses spline regression, however without identifiability conditions.
Methods based on
regression error are able to decide between Markov equivalent
DAGs under the assumption of having a non-linear function
and low noise \cite{bloebaum2018}. The method we present in this paper will also adopt these assumptions.


In this paper, we  propose a  causal score using the Minimum Description Length (MDL) of regression models via functions from  dense functional classes.  
A functional class is dense in the set of all continuous functions $C([a,b])$ on a compact interval $[a,b]$ if it 
can approximate any continuous function on   $[a,b]$ within arbitrary precision.  Under certain
assumptions on the causal models, we prove identifiability of
these causal scores using MDL of functions from dense classes. 
We propose method LCUBE as an  instantiation of  this MDL-based score  for  the  cubic regression splines which possess the density property.  LCUBE uses MDL encoding  that allows more detailed
characterization of cubic splines models than the encoding via
AIC or BIC criteria.

\vspace{0.3cm}
The key contributions of this work can be
outlined as follows: 

\begin{itemize}
 \item  We introduce a bivariate causal score based on the  MDL principle, using functions that possess the density property on a compact real interval. We prove the identifiability of these causal scores under specific conditions, which can be easily tested in practice.
 \item  We propose LCUBE as
an instantiation of the MDL-based causal score using cubic
regression splines. LCUBE is an identifiable method that is also
interpretable, simple and very fast.
 \item Empirical evaluations
compared to state-of-the-art methods demonstrate  that LCUBE achieves superior precision in terms of AUDRC on the real-world Tübingen cause-effect pairs dataset. 
 It also outperforms  the methods in terms of average precision across 10 common benchmark datasets and achieves above-average precision on  13 datasets.
\end{itemize}

\section{Preliminaries}

Recent literature on causal inference imposes
 certain model restrictions in order to to identify the cause and effect reliably. For example, \cite{shimizu2006linear}, \cite{peters2012identifiability}  provide   theoretical foundations to establish identifiability  of non-linear additive noise models. 
In our work, we assume that  causal relations between $X$ and $Y$ can be expressed by  a non-linear  bivariate functional model, an instance of the Additive
Noise Models (ANM), \cite{shimizu2006linear},  and that there are   no other causes of $Y.$ 
Our models  further rely on   the  assumptions on regression  from \cite{bloebaum2018}.
 Blöbaum et al. in \cite{bloebaum2018} formulated the following three  assumptions under
which the true causal direction is identifiable via regression.

\begin{assumption}\cite{bloebaum2018}
     (Causal model). The effect can be written as

\begin{equation}\label{ANM_alpha}
Y_\alpha := f(X) + \alpha N,
\end{equation}
with noise term \(N\) and parameter \(\alpha\) restricting the noise level.
\end{assumption}

\begin{assumption} \cite{bloebaum2018} 
\label{Assumption2}(Unbiased noise). The noise term \(N\) is unbiased and has unit variance.
\end{assumption}

\begin{assumption} \cite{bloebaum2018}
 (Compact supports). The distribution of \(X\) has compact support and w.l.o.g. \(X\) attains values between 0 and 1, which can be achieved by normalizing \(X\). Further, the distribution of \(N\) has compact support and there exist values \(r_+ > 0 > r_-\) such that for each value \(x \in X\), \([r_-, r_+]\) is the smallest interval containing the support for the conditional density of \(N\) given \(x\). Hence, we know that \([ \alpha r_-, 1 + \alpha r_+ ]\) is the smallest interval containing the support of the density of \(Y_\alpha\) and rescale it to

\[
\tilde{Y}_\alpha := \frac{Y_\alpha - \alpha r_-}{1 + \alpha r_+ - \alpha r_-}.
\]
\(\tilde{Y}_\alpha\) has the same scale as \(X\) and has values between 0 and 1.
\end{assumption}

\noindent Now  consider an integer $n \ge 2$ integer  and  $\alpha:= \frac{1}{n}$ Eq.~(\ref{ANM_alpha}). With a slight abuse of notation, we write $Y_n$ instead of $Y_{\frac{1}{n}}$, and define the following  ANM model  

\begin{equation}\label{ANM_alpha2}
Y_n := f(X) + \frac{1}{n} N
\end{equation}
with noise  \(N\). One can see that   (\ref{ANM_alpha2}) is a special case of (\ref{ANM_alpha}).
We stress here that the noise variable $N$ in Eq.~(\ref{ANM_alpha}).
does not have to be necessarily Gaussian.

\subsection{Identifiability of likelihood score-based methods in causal discovery}

Schultheiss and Bühlmann in \cite{schultheiss}
 discuss causal discovery in structural causal models using Gaussian likelihood scoring and analyze the
effect of model misspecification in a general directed acyclic graphs.
They came to the following conclusions in the case where the data-generating distribution comes from a linear structural equation model and
linear regression functions are used for estimation. When the true error distribution is
Gaussian, one can only identify the Markov equivalence class of the underlying data-generating DAG. The
same holds true when the error distribution is non-Gaussian, but one wrongly relies on a Gaussian error
distribution for estimation.
When likelihood scores in a linear structural equation model and linear regression are used for comparison of different DAGS and  Gaussianity of errors is not assumed,  the multivariate density of the DAGs factorize only for independent error functions. 
The true DAG maximizes this likelihood score due to the properties of the Kullback-Leibler
divergence. In practice, this comes with the additional difficulty of estimating the densities  and one
needs to add some penalization to prefer simpler graphs and avoid selecting complete graphs \cite{nowzohour}.

In this paper, we face the pitfalls associated with likelihood-based causal scores by employing minimum description length 
(MDL)-based scores.
The MDL-based scores  inherently penalize more complex graph structures and do not assume Gaussianity of the error terms. 
 As we demonstrate in our experiments using concrete MDL scores, this approach yields high causal accuracy in practice.
 
\section{Functional model with density property and MDL}


\subsection{Density property of a  functional class}

In general, a subset $A$ of a topological space $X$ is said to be dense in $X$ if every point of $X$ either belongs to $A$ or is arbitrarily "close" to  a point in $A$. In functional approximation theory, the Weierstrass Approximation Theorem states that any given complex-valued continuous function defined on a closed interval $[a,b]$  can be uniformly approximated as closely as desired by a polynomial function. 
In other words,  polynomial functions are dense in the space   of continuous complex-valued functions on  $[a,b]$, when equipped with the supremum norm. 
For the purposes of our paper we restrict ourselves  to real-valued continuous functions  on a closed real interval 
$[a,b]$, which is denoted  as 
$C([a,b])$.

Assume now a functional causal model from Eq.~(\ref{ANM_alpha2}) with error $\frac{1}{2n}N$ where we
replace $f$ by its dense approximation $h$ within error 
$\frac{1}{2n}N$ (for the supremum norm on $C[a,b]$). 
So for a sufficiently large $n$,
we get from 
Eq.~(\ref{ANM_alpha2}) model
\begin{equation}\label{ANM_alpha3}
Y_n \approx h(X) + \frac{1}{n}N.
\end{equation}

Since $Y_n$ can be approximated by $h$ on the compact interval $[a,b]$ within error $\frac{1}{n}N,$  the guarantee of this error  holds also for every subset of $y_i=f(x_i), i=1, \dots, n$, drawn from continuous  $X$  on a compact interval.
Therefore given an  approximation of $f$ by $h$ from the dense functional class $H$, we can express the regression model  on finite samples 
as

\begin{equation}\label{ourANM2}
y_i \approx f(x_i)  \approx h(x_i, \btheta), \hspace{0.5cm} i=1, \dots, n.
\end{equation}

\subsection{Kolmogorov complexity and MDL}

 To be able to
distinguish between Markov equivalence classes, Janzing and
Schölkopf in \cite{janzing2010}  postulated the algorithmic equivalent of
the principle of independent mechanisms.
Under the algorithmic model
of causality, they formulated the postulate of algorithmic independence of conditionals using Kolmogorov complexity.  This led to the following inference rule, as described   in \cite{stegle}:
If $X \to Y$ is the true causal direction, then
for Kolmogorov complexity $K$ of a cause $X$ and effect $Y$ given cause holds

\begin{equation}\label{rule_by_KC}
K(P(X)) + K(P(Y|X)) \overset{+}{\le } K(P(Y)) + K(P(X|Y)) 
\end{equation}
where $\overset{+}{\le }$ denotes inequality up to additive constants.
However, the Kolmogorov complexity is not computable \cite{li_vitanyi} and   the true distribution is not known, 
 only a  data sample is usually available. A principled way to
solve at least the first part of the problem is to approximate
Kolmogorov complexity via the Minimum Description Length
principle (MDL). The Minimum Description Length principle
is a practical variant of Kolmogorov complexity that approximates $K$ from above,  see e.g.,  \cite{rissanen}, \cite{grunwald1998minimum}.
The MDL principle has been applied in  bivariate causal discovery for classes of probability functions, as in e.g. \cite{buthathoki}, \cite{formally}.
 Given
data $D$, which may represent a sample from a distribution
$P(X)$, the idea is to find that model $M^*\in {\cal M}$, such that

\begin{equation}\label{MDL_for_functions} 
M^*=   argmin_{M\in {\cal M}} L(M) + L(D|M)
\end{equation}

where $L(M)$ is the length in bits  required  to describe the model
$M$  within the model class ${\cal M}$, and $L(D  |M)$ is
the length in bits of the description of data $D$ given $M$. 
For  more
details to MDL and its coding, we refer the reader to  \cite{grunwald1998minimum}.

In this paper
we
consider the model class ${\cal M}$ to be
a  class of functional models.
We assume that each $f$ in the functional class ${\cal M}$ 
   can be parameterized by a  vector
$\btheta$ and each function $f$ can be fitted to the data sample $D$.  Accordingly,  in (\ref{MDL_for_functions}), $L(M)$ measures the complexity of describing the parameter vector $\btheta$, and $L(D | M)$
quantifies the goodness of fit of the function $f$ given by the
parameter vector $\btheta$ to on the data $D$. 
In this way, the causal direction rule from Eq.~(\ref{rule_by_KC}) becomes

\begin{equation}\label{rule_by_MDL}
\min_{\btheta_X} (L(X) + L(Y|X, \btheta_X)) \overset{+}{\le } \min_{\btheta_Y} (L(Y) + L(X|Y, \btheta_Y)) 
\end{equation}
where $\btheta_X, \btheta_Y$ are parameterizations of each  causal model direction.
If we  assume  for cause that it is  normalized  and has a uniform prior, 
then the codes $L(X),L(Y)$  do not depend on the parameterizations $\btheta_X, \btheta_Y$,  
and we can instead use the rule
\begin{equation}\label{rule_by_M2DL}
\min_{\btheta_X}  L(Y|X, \btheta_X) \overset{+}{\le } \min _{\btheta_Y} L(X|Y, \btheta_Y). 
\end{equation}

 We use  the mean residual sum of squares (MRSS) as the goodness of fit. For the  expected least-squared error $E(.)$, we will use the commonly adopted abbreviation ELSE. 

\section{
Identifiability of the MDL Causal Score via Dense Functional Classes}

Let  $\beta_f$ denote the set of parameters of a function $f$ and let
 $\|\beta_f \|_0$  denote its number of non-zero parameters. 
 We will further assume  for cause that it is  normalized  and has a uniform prior.  

\begin{assumption}
 \cite{regularizedregression} (Simplicity). Let \(Y_\alpha\) be generated as in Assumption 1.  Let \(\varphi\) be the function minimizing the ELSE for predicting the effect \(Y\) from the cause \(X\) and \(\psi\) be the function minimizing the ELSE in the anti-causal direction. We assume that \(\psi\) has at least as many parameters as \(\varphi\), i.e. \(\| \beta_\varphi \|_0 \leq \| \beta_\psi \|_0\).
\end{assumption}

And now we can present our result:
\begin{theorem}\label{th0}
 Let $H$ be a dense functional model class  where each function from $H$ is   parameterized by vector $\btheta$. Assume the Assumptions 1-4 for causal models given by Eq.~(\ref{ANM_alpha3})  between cause and effect are satisfied.  
 Let the encoding of the parameter vector 
 $\btheta$
 in the MDL description, for a sample set of size 
$n$, be given by the logarithm of a positive integer function that is strictly increasing with 
$n$. 
Assume that the goodness-of-fit measure is either ELSE  or MRSS. Then, for the score function 
   $L(\tilde{Y}_n|X, \hat{\btheta}_X)$ from Eq.~(\ref{rule_by_M2DL})  to infer the causal direction ${X\to Y}$, holds
 
\begin{equation}\label{limit1}
\lim_{n \to \infty} \frac{L(\tilde{Y}_n|X, \hat{\btheta}_X) }{L(X|\tilde{Y}_n, \hat{\btheta}_Y)} \le 1,
\end{equation}

with equality if and only if  the function $f$ in Eq.~(\ref{ANM_alpha3}) is  linear.
\end{theorem}
\emph{Proof:}
To prove the theorem, we need the following definition  from \cite{regularizedregression} and our  Lemma 1.
\begin{definition}\label{IRSFdef}
     (Identifiable Regression-based Scoring Functions) \cite{regularizedregression}. Given two random variables \(X\) and \(Y\) and a regression function \(\varphi\) that maps \(X\) to \(Y\), and  a scoring function \(S : \mathbb{R}_{\geq 0} \times \mathbb{N} \to \mathbb{R}\)   which takes as input the expected least-squared error \(E[(Y - \varphi(X))^2]\) and the number of parameters of \(\varphi\), \(\| \beta_\varphi \|_0\). A scoring function
$$
S(Y \mid X, \varphi) := \gamma (E[(Y - \varphi(X))^2]) + \lambda(\| \beta_\varphi \|_0)
$$
is called an Identifiable Regression-based Scoring Function (IRSF), if both \(\gamma : \mathbb{R}_{\geq 0} \to \mathbb{R}\) and \(\lambda : \mathbb{N} \to \mathbb{R}\) are strictly  increasing.
\end{definition}

\begin{lemma}\label{lemmaKS}
Let  the causal model  be given by 
Eq.~(\ref{ANM_alpha3})
where $Y_n$ denotes the effect $Y$ estimated via regression using  $n$ samples. Assume that Assumptions 2–4 hold for this model. Let
      \(\varphi\) denote the function that minimizes the ELSE when predicting the effect \(Y\) from the cause \(X\),  and let
      \(\psi\) be the function minimizing the ELSE for predicting \(X\) from \(Y\), i.e. \(\varphi(x) = E[Y \mid x]\) and  \(\psi(y) = E[X \mid y]\). Let \(S\) be an IRSF as defined in Definition 1. Then the following limit always holds
\begin{equation}\label{IRSFequation}
\lim_{n \to \infty} \frac{S(E[(\tilde{Y}_n - \varphi(X))^2], \| \beta_\varphi \|_0)}{S(E[(X - \psi(\tilde{Y}_n))^2], \| \beta_\psi \|_0)} \le 1,
\end{equation}
with equality if and only if \(\varphi\) is linear.
\end{lemma}
\emph{Proof:}
We have moved the proof of Lemma 1 to Appendix~\ref{proofL1} due to its technical nature.
 We note that a similar lemma was proven as Theorem 1 in \cite{regularizedregression}, although for the model given by
  Eq.~(\ref {ANM_alpha}) and $\alpha \to 0$.
Regarding our Theorem 1, we will show that  $L(\tilde{Y}_n|X, \hat{\btheta}_X)$ satisfies the conditions of Lemma~\ref{lemmaKS}. 
In the nominator an denominator of 
Eq.~(\ref{limit1})  the parameterizations $\hat{\btheta}_X, \hat{\btheta}_Y$ are fixed, and therefore do not affect  the limit in Eq.
~(\ref{limit1}).
The  fulfillment of the IRSF conditions for  the function $\lambda: \mathbb{N} \to \mathbb{R}$  is straightforward:
Since  $b(n)$ is  strictly increasing in $n$,  it follows that also 
$\log(b(n))$   is strictly increasing, thus it  holds for
\(\lambda : \mathbb{N} \to \mathbb{R}\), too.
Secondly we show the fulfillment of the conditions for the function \(\gamma : \mathbb{R}_{\geq 0} \to \mathbb{R}\).
The statement is obvious for the goodness of fit  ELSE, since $\log$ is a strictly increasing function.  For  MRSS:
In a discrete finite scenario, 
MRSS reflects  empirical variations but aligns with ELSE for large $n$. As the number of data samples $n$ 
 converges to infinity, MRSS converges to ELSE, assuming unbiased estimation, see  
 \cite{vapnik}. Functional spaces with the density property on a compact interval provide unbiased estimation for a sufficiently large $n$.
So for sufficiently large $n$ we can approximate

\begin{equation}\label{approx1}   
E[(\tilde{Y}_{n} - \varphi(X))^2] \approx  \frac{RSS(\hat{\btheta}_X)}{n}
\end{equation}

\begin{equation}\label{approx2}
\mbox{ and } E[(X - \psi(\tilde{Y}_{n}))^2] \approx \frac{RSS(\hat{\btheta}_Y)}{n}
\end{equation}

\noindent with $\hat{\btheta}_X$, $\hat{\btheta}_Y$  parameterizations of models corresponding to $X\to Y$ and $Y\to X$, respectively. 
Thus for function  $\gamma:=\frac{n}{2} \log \frac{RSS(\hat{\btheta}_X)}{n}$ holds $\gamma: \mathbb{R}_{\geq 0} \to \mathbb{R}$ and this function is strictly  increasing   for $n$.
Therefore the MDL score is IRSF according to Definition 1 and the statement of Theorem 1 follows.
\qed

\noindent In the following,  we  use our instantiation of the cubic spline functional class for the set $H$. This class  has  the density property on $C([0,1])$  and we   construct the MDL-based  causal score about  which we show that it is   IRSF.  In this way, the  identifiability of the score under the given mild conditions is proven.

\section{Cubic spline regression for bivariate causal discovery}\label{csr_for_biv_cd}

\subsection{The class of cubic splines and its density property}\label{csr_for_biv_cd_sub}

Cubic spline functions have been well studied over decades as they  provide reliable approximation   of continuous functions.
The class of cubic splines with variable number of knots is dense, i.e.,
  any   continuous function having fourth derivative on a compact interval can be approximated by a  cubic  spline within an arbitrary precision, having  an arbitrary finite number of knots, \cite{schultz1973spline}.
However, the parametric representation  of any continuous function by cubic spline does not have to be unique.
 The density property holds even more general for degree $r \ge 3$
 as we sketch in the following.
In general, on a compact interval, for the subclass of continuous functions, splines of best approximation do have uniqueness properties in the case when they are tied together from smooth functions to form a Chebyshev system on $[a,b]$ 
\cite{galkin1974uniqueness}. 
\cite{subbotin1970order}.  Theorem 3  in \cite{subbotin1970order} shows that if one approximates a function of smoothness $r-1$  by splines $s$ of degree $r$ with variable knots and the number at most $m$, then  at the expense of the choice of knots, the order  of approximation, i.e. $\|f-s\|_{L_p[a,b]}$  is equal to $m^{-r-1+i}$ where $i \le r-1$ is integer and $m$ is the number of knots and $f:[a,b] \to \mathbb{R}$ is a given  continuous function.
  Based on the above results of Subbotin in \cite{subbotin1970order}, with the increasing number of knots $m$,   a continuous function can be approximated by a cubic spline function within  arbitrary precision.
For interpolation on a finite set, 
 spline interpolant exists and is unique for a given set of data points, provided the knots and interpolation points satisfy certain geometric conditions, see  the Schoenberg-Whitney conditions \cite{schultz1973spline}.

 Consider a curve to be a mapping associated by   parameters
$\beta_j$  with a
fixed degree and fixed knots.
A  spline  of degree $r$ with knots at $k_j,j=1, \dots,m$ is a piecewise  polynomial with continuous derivatives up to order $r-1$ at each knot.
If the degree and the knots are fixed, then we have a vector space of piecewise polynomials. The B-spline functions form a basis for this vector space; Thus the coefficients $\beta_j$  of any given curve in this vector space are uniquely determined \cite{de1978practical}.

\subsection{Representation of $(X,Y)$ by a cubic regression spline}\label{Bsplines}

We will model the variable 
 $Y$  as a  cubic spline function  of the variable $X$, both of them given by datasets and affected by  independent
noise. The  spline is given by the dataset and by the parametrization  as in Eq.~(\ref{ourANM2}) and we  will  approximate $f$  
by a cubic regression spline with $m$ knots
\vspace{-0.2cm}
\begin{equation}\label{Eq:regression_spline}
f(x) \approx  s_f(x) = b_0 + b_1x +  b_2 x^2 +  b_3 x^3 +  \sum_{j=1}^m \beta_j(x-k_j)_+^3.
\end{equation}

\noindent $k_j$ is the position of the j-th knot, $\{b_0, b_1, b_2,, b_3, \beta_1, \dots, \beta_m\}$ is the set of coefficients and $(c)_+ = \max (0,c)$.
We further assume that 
$\min(x_i) < k_1 < \dots < k_m < \max(x_i)$ and $\{k_1, \dots, k_m\} \subset \{x_1, \dots, x_n\}$. Denote $\bf{k}$ $ = (k_1, \dots, k_m)^{\top}$, $\bb = (b_0, b_1, b_2, b_3)^{\top}$ and 
$\bbeta = (\beta_1, \dots, \beta_m)^{\top}$.
If function $f$ can be approximated by the spline function $s_f$ from (\ref{Eq:regression_spline}), then the estimate of  $f$ can be obtained by estimating $m,  \bf{k}$, $\bf{b}$ and 
$\bf{\bbeta}$. Thus the problem of estimation of $f$ becomes a model selection  problem.
We restrict ourselves to cubic splines  with  equidistant knot sequences, since  
this class also has the density property in a compact interval, see e.g.\cite{wahba}.
\subsection{Cubic spline parametric approximation of a continuous function}

In the following we will construct the MDL  description of a cubic 
regression spline function.
The problem of estimating  the function $s_f$ in (\ref{Eq:regression_spline}) 
can be seen as the model selection problem  where the model is characterized by a parameter vector

\begin{equation}\label{omega}
\btheta =(m, \bk, \bb, \bbeta)
\end{equation}
where  $m$ the number of knots, $\bb, \bbeta$ the vectors of coefficients and $\bk$ the vector of $m$ knot points.
Assume  now the estimates $\hat{m}, \hat{\bk}$ are known.
Denote vectors $\bx =(x_1, \dots, x_n)^{\top}$, $\by =(y_1, \dots, y_n)^{\top}$ and design matrix
\begin{equation}\label{design matrix}
\bX :=({\bf 1}, \bx, \bx^2, \bx^{3}, (\bx -k_1 {\bf 1})^{3}_+, \dots,(\bx -k_{m} {\bf 1})^{3}_+)
\end{equation}
where ${\bf 1}$ is $n \times 1$ vector of ones.
Then if $m, \bk $, are already specified, by using normal equations for multivariate regression we can get the estimates  of $\bb, \bbeta$ as

\begin{equation}\label{Eq:beta_estimate}
(\hat{\bb}^{\top}, \hat{\bbeta}^{\top} )= (\bX^{\top} \bX) ^{+} \bX^{\top}
\by,
\end{equation}
where $A^+$ is the inverse or pseudo-inverse  of matrix $A$. 
(Since knots are not fixed, we do not have to find matrix $\beta$ uniquely, therefore pseudoinverse.)
Eq.~(\ref{Eq:beta_estimate}) is the least squares estimate of $\bf{b}, \bbeta$ given estimates of $m$, $\bk$.
If for  the error term holds $N \sim N(0,1)$ in Eq.~(\ref{ANM_alpha}),  the estimate of $\bf{b}, \bbeta$ in Eq.~(\ref{Eq:beta_estimate}) is also the maximum likelihood estimate.

\subsection{Causal rule by cubic spline approximation}

We will compute our scoring function  by MDL and denote it for both causal   directions    as $\hat{\Delta}({X \to Y})$ and 
$\hat{\Delta}({Y \to X})$. 
To be able to compute these scores,    we need to compute 
$L(Y|X,\btheta)$ and $L(X|Y, \btheta)$.
$L(Y|X, \btheta), L(X|Y, \btheta)$ will be computed as the MDL two-part code for the cubic spline regression functions $\hat{s}_f$ and $\hat{s}_g$, where $f$ and $g$ are functions for each causal directions $X \to Y$ and $Y \to X$, respectively,  as in  Eq.~(\ref{Eq:regression_spline}).  The code lengths of $L(Y|X, \btheta), L(X|Y, \btheta)$  are explained in the following section.

\subsection{Code length   $L(Y|X, \btheta)$ and  $L(X|Y, \btheta)$}\label{code_length_coditionals}

\noindent Now we will compute codes $L(Y|X, \btheta)$ and $L(X|Y, \btheta)$.
With a slight abuse of notation and for simplicity we do not write $\tilde{Y}_n$ but     $Y= \{y_1, \dots, y_n\}$ for a cubic spline regression  on set
$X=\{x_1, \dots, x_n \}$, i.e.

\begin{equation}
Y \approx  s_f(X)
\end{equation} 
where $s_f$ is  defined by (\ref{Eq:regression_spline})  with $m$ knots and analogously

\begin{equation}
X \approx  s_g(Y)
\end{equation} 
where $s_g$ a cubic spline function for the opposite direction.
Without  the lack of generality, we explain only $L(Y|X)$.
The estimation  of $Y$ by  $s_f(X)$ is a model selection problem
where each model is defined by parameter 
$\btheta$.
Having parameter vector  $\btheta = (m, \bk, \bf{b}, \bbeta)$  from Eq.~(\ref{omega})  corresponding to a fitted spline model, 
we  compute  the two-part MDL code 
as
\begin{equation}\label{two_part_MDL}
L(Y|X, \btheta) := 
 L(\btheta) + L(Y|\btheta). 
\end{equation}
$L(\btheta)$ can be still decomposed  into


\begin{equation}
L(\btheta) =    L(m) + L(\bk |m) + L(\bf{b}, \bbeta, |  m,\bk)
\end{equation}
and $L(Y|\btheta)$ is a goodness of fit.  
We derive 

  \begin{equation}\label{Ltheta}
  L(\btheta) = \log m   
+ \sum_{j=1}^{m} \log u_j + \frac{m+ 4}{2} \log n 
\end{equation}
where $u_j$ are defined as follows:
Let $k_0 = \min(x_i)$
and $k_{m+1} = \max (x_i)$
and similarly as in \cite{lee2000regression} let

 \begin{equation}\label{u_j}
 u_j :=  \mbox{the  number of } x_i's \mbox{ so that }  k_{j-1} \le x_i \le k_j
 \end{equation}
for $j=1, \dots, m$. 
Each $u_j$ is the $j-th$ successive (i.e. next) index difference.  If we know the whole sequence $(u_1, \dots, u_{m})$,  we also know the whole knot sequence $\bk$. 
For fluency  in  reading we moved the full derivation of $L(\btheta)$  to Appendix~\ref{derivation_Ltheta}.

\subsection{Code length  $L(Y|\btheta)$}
Now we code the goodness of fit, using the code from \cite{rissanen2007stochastic}, as the code of $Y$ given $\btheta$ by 
\begin{equation}\label{codelenght_by_RSS}
L(Y| \btheta) = \frac{n}{2} \log (\frac{RSS(\btheta)}{n}) + C
\end{equation}
where the residual sum of squares  $RSS(\btheta) := \sum_{i=1}^n (y_i - s_f(x_i| \btheta))^2$ and C is a  constant which is for all models the same and thus we  can be omitted  from MDL  for the model selection problem.

\subsection{The full code $L(Y|X, \btheta)$}\label{final_code}

\noindent The full code for conditionals is

\begin{align}\label{case_r1}
L(Y|X, \btheta) = L(\btheta) + L(Y|\btheta)   = \log m   
+ \sum_{j=1}^{m} \log u_j + \frac{m + 4}{2} \log n  + 
\frac{n}{2} \log (\frac{RSS(\btheta)}{n}) 
\end{align}
\noindent where  $RSS(\btheta) = \sum_{i=1}^n (y_i - s_f^{m,\bk}(x_i))^2$ and $s_f^{m,\bk}(x_i)$ is a  cubic regression spline  with knot sequence $\bk$ with $m$ knots.
Algorithm~\ref{alg_L_theta_7} summarizes these computations.

\begin{algorithm}[H]
\caption{$\hat{\Delta}_n({X\to Y})$ by cubic spline $s_f$}
\label{alg_L_theta_7}
\begin{algorithmic}[1]  
  \STATE \textbf{Input:} $X$, $Y$ given by $(x_i, y_i)$, $i = 1, \dots, n$; integer $m_{\text{max}}$
  \FOR{$m \le m_{\text{max}}$}
    \STATE Compute equidistant knot sequence $\bk$
    \STATE Compute values $\bb, \bbeta$ from Eq.~(\ref{Eq:beta_estimate})
    \STATE Compute $u_j$ values from Eq.~(\ref{u_j})
    \STATE Compute $L(Y|X, \btheta)$ from Eq.~(\ref{case_r1}) for $\btheta = (m, \bk, \bb, \bbeta)$
  \ENDFOR
  \STATE Compute $\hat{\Delta}_n({X\to Y}) := \min_{\btheta} L(Y|X,\btheta)$
  \STATE \textbf{Output:} $\hat{\Delta}_n({X \to Y})$
\end{algorithmic}
\end{algorithm}

\begin{algorithm}[H]
\caption{Decide causal direction}
\label{alg_causal_score}
\begin{algorithmic}[1]  
    \STATE \textbf{Input:} $\hat{\Delta}_n(X \to Y)$ and $\hat{\Delta}_n(Y \to X)$ from Algorithm~\ref{alg_L_theta_7}
    
    \IF{$\hat{\Delta}_n(X \to Y) < \hat{\Delta}_n(Y \to X)$}
        \STATE $CD := X \to Y$
    \ELSIF{$\hat{\Delta}_n(X \to Y) > \hat{\Delta}_n(Y \to X)$}
        \STATE $CD := Y \to X$
    \ELSE
        \STATE $CD$ is undecided
    \ENDIF
    
    \STATE \textbf{Output:} Causal direction $CD$
\end{algorithmic}
\end{algorithm}

\noindent To compute  $\hat{\Delta}_n({X \to Y})$ in Algorithm~\ref{alg_L_theta_7}, we take  the  maximum number of equidistant knots $m_{max}$ of the cubic regression splines as a hyperparameter.
Based on the observation from \cite{subbotin1970order} mentioned in Section~\ref{csr_for_biv_cd_sub},  the parameter $m$ controls the precision of the uniform approximation of  the function $f$ by  the spline $s_f$.
Algorithm~\ref{alg_causal_score} determines the causal direction between the scores 
$\hat{\Delta}_n(X \to Y)$ and $\hat{\Delta}_n(Y \to X)$.

\subsection{Identifiability of the MDL cubic  spline regression score}

\begin{corollary}\label{th2}
Assume that the causal direction $X \to Y$ for  the  dataset $(X,Y)$ with $n$ samples  is modeled by Eq.~ (\ref{ANM_alpha3})  and the score $\hat{\Delta}_n({X\to Y})$ was computed by  Algorithm~\ref{alg_L_theta_7} for even $m$.
 Then 
 \[
\lim_{n \to \infty} \frac{\hat{\Delta}_n({X\to Y})}{\hat{\Delta}_n({Y\to X})} \le 1,
\]
with equality if and only if  the function $h$ in Eq.~(\ref{ANM_alpha3}) is linear.
\end{corollary}
\emph{Proof:}
We prove Corollary~\ref{th2} by showing that $L(Y|X, \btheta)$  code from Eq.~(\ref{case_r1}) is IRSF according to Definition 1.
Assumption 1 is satisfied, since  both Eq.~(\ref{ANM_alpha2}) and Eq.~(\ref{ANM_alpha3})  are  special cases of Eq.~(\ref{ANM_alpha}). 
Assumption 2 is  satisfied  for both models from Eq.~(\ref{ANM_alpha2}) and Eq.~(\ref{ANM_alpha3}) due to the density of the class of cubic splines in $C([0,1])$ with an arbitrary number of knots $m$.
Assumption 3:   Using spline regression, compact supports are assumed for numerical stability.
Assumption 4:
It holds  that if we know that $\varphi$ from Definition 1 consists of a linear combination
of basis functions that are linearly independent of each other, we
cannot find an inverse function that has fewer degrees of freedom, see \cite{regularizedregression}.
Concretely,  Kilbertus et al. in \cite{kilbertus} showed that for low degree polynomials, 
 it is not possible to formulate an inverse with less parameters as
the original function. 
We can therefore assume that  \(\| \beta_\varphi \|_0 \leq \| \beta_\psi \|_0\), i.e.   Assumption 4 is satisfied for  the MDL cubic spline regression score.
In Definition 1, if $\varphi$ is the function minimizing
the expected least-squared error (ELSE) for predicting the effect $Y$
from the cause $X$, then in case of MDL cubic regression score for a fixed $m$, it holds   $\varphi := \hat{s}_f$ and if \(\psi\) is the function minimizing the ELSE in the anti-causal direction, which is in case of MDL cubic spline regression score for a fixed $m$  then it holds $\psi := \hat{s}_g$. 
We define

\begin{align}
\lambda(\| \beta_\varphi \|_0) :=   \log m    + \sum_{j=1}^{m} \log u_j^X   +  \frac{m + 4}{2} \log n = \log ( m \times \prod_{j=1}^m u_j^X \times n^{\frac{m + 4}{2}}).
\end{align}

The value $m$ depends on $n$. The product
$m  \times \prod_{j=1}^m u_j^X  \times n^{\frac{m + 4}{2}}$  has for even $m$ values in  $\mathbb{N}$
 and thus for $\lambda:= \log(...)$  holds that  \(\lambda : \mathbb{N} \to \mathbb{R}\); Function $\lambda$ is strictly  increasing function for $n$.
Analogically holds the same  also for  
$ \lambda(\| \beta_\psi \|_0)$
for the direction  $Y \to X$.
The relationship between ELSE and MRSS  holds also for the class of cubic splines, as this  class  is an instance of a dense functional class in  Theorem~\ref{th0}.
As the number of knots $m$ increases and the data sample size $n$  grows, due to the density property of cubic splines,  the MRSS  converges to ELSE.
\qed

Our method for computation of causal score  which is performed by Algorithm~\ref{alg_L_theta_7} and 
Algorithm~\ref{alg_causal_score} is called  LCUBE, as an abbreviation for mdL CUBic splinE score.

\vspace{0.3cm}

\noindent \textbf{Complexity of the LCUBE algorithm:} The complexity of computing the regression spline $s_f^m$ in Algorithm~\ref{alg_L_theta_7} is $O(n m + m^3)$. The complexity of LCUBE by Algorithm 1 for $m \le m_{max}$ together with Algorithm~\ref{alg_causal_score}  is $O(2 m_{max}(n m_{max} + m_{max}^3))$. Further details on the computation are provided in Appendix~\ref{complexityLCUBE}. In our experiments,  $m_{max}$ was at most 10.

\subsection{Differences of LCUBE to Slope and Sloppy}

There are two methods,  Slope \cite{marx2017telling}
and Sloppy \cite{regularizedregression},
which  use spline regression, and therefore are related to LCUBE. Slope employs a 
two-part MDL to approximate the algorithmic Markov condition
for continuous data. 
It also assumes  that the error is Gaussian
distributed.  The score
used in Slope can be written as $\gamma (E[(\tilde{Y}_{\alpha} - \varphi(X ))^2])$,where $\gamma$ is
based on the negative log likelihood, plus a function $\rho$ over the
parameters. Since $\rho$ does not purely consider the number
of parameters, but assigns different weights according to the value,
of the parameter, 
 the corresponding scoring function is
not an IRSF. No identifiability results are known for Slope. 
Sloppy is 
 is the first instantiation
of the IRSF framework. 
Method Sloppy (Version Sloppy$_S$) fits a cubic spline, where  the degrees
of freedom is controlled and that selection of splines is found, for which S
is minimal. As a scoring function $S$, AIC and BIC are used.

The scoring function of LCUBE is the $L(Y|X,\btheta)$ score from Theorem~\ref{th0} and it satisfies Assumptions 1-4 for cubic splines. 
Moreover, our MDL encoding   allows more detailed characterization of cubic spline models than the encoding via AIC or BIC applied by Sloppy.

\section{Approximation by  dense functional classes and  identifiability of the  LCUBE score}

\subsection{The classes of cubic splines and  harmonic functions and their density property}

For some functional classes  that possess the density property, the upper bounds on the approximation errors   are known.
In approximation theory, the rate of approximation  on a compact interval describes how quickly an approximating function converges to the target function as the approximation parameters, such as polynomial degree or number of basis functions,  improve. 
The density property of cubic splines was discussed in Section~\ref{csr_for_biv_cd_sub}. A well known upper bound  from \cite{hall}  states that the class of cubic splines approximates smooth functions with an error rate of $O(h^4)$ where $h$
is the spacing between knots. 
 In Fourier approximation,  the approximation error  depends on the function's smoothness and  decreases at a rate of  order  $O(1/n^2 )$ for differentiable functions, where $n$ is the number of terms  included in the Fourier series expansion, see \cite{kovachki}.
In area of neural network approximation, 
the rates of approximation refer to how efficiently neural networks can approximate functions. These are  measured in terms of the number of neurons or layers required to achieve a certain level of accuracy. For an overview, see e.g. \cite{ratesKHS}.
Regarding  cubic splines with the  equidistant knot sequence $\bk$ and  values $\bb, \bbeta$, we assume that they  satisfy the Schoenberg-Whitney conditions \cite{schultz1973spline}. These  ensure   the existence of a unique cubic spline that interpolates the given data points.
As  we will see from the experiments with LCUBE in Section~\ref{exp_res}, these properties of cubic splines 
contribute to its high causal score, particularly for datasets with low noise.


 We showed above that the set of cubic splines satisfies  Assumption 4. Intuitively, if $X$ causes $Y$, the functional causal model $Y=f(X)$
 may exhibit a smoother relationship compared to the anti-causal model $X=g(Y)$.
 This could mean that
the functional dependence of $Y$
 on $X$
 is more regular and smoother, potentially requiring fewer knots in spline approximation than for the functional dependence of $X$ on $Y.$ The mapping from $Y$ to $X$
 may be more complex, e.g., nonlinear or multi-valued, thus requiring more  spline knots to adequately capture  its variations.

\subsection{The class of shallow   feed-forward neural networks and its density property}\label{density_of_FNN}

 \noindent The class of feed-forward neural networks  with one hidden layer and $d$ input units defined on $C([a,b]^d)$ (i.e., shallow feed-forward networks),    has the universal approximation property \cite{hornik1989multilayer}. This property  states that for any sufficiently smooth function 
$f$ on a compact set with finitely many discontinuities, there exists a feedforward network with one hidden layer  (abbreviated as FNN), denoted $F$ 
that can approximate $f$ arbitrarily well if 
the number of hidden units  $m$
in $F$ is sufficiently large and
the activation function  $\sigma$ is not polynomial. In other words, the FNN  class is dense in $C([a,b]^d)$.
The literature provides upper bounds on the number of hidden units required to achieve a given approximation error on a compact interval for various activation functions, e.g. \cite{kainen}, \cite{hlavackova}. 
These upper bounds can serve  as an upper bound on $m$ while keeping the the approximation error in a compact interval under some value. However, in a finite sample scenarios, these bounds are  too high in practice.
However, directly constructing a FNN and estimating its approximation error on a finite dataset is challenging.
  Such a FNN would have the generalization ability without being trained. 
 It is well known that the output of an FNN depends heavily on the chosen training algorithm and its parameter settings. In short, to find encodings that characterize the structure and learning  of FNNs for both causal and anti-causal models, such that they satisfy Assumption 4, is a difficult task.

\section{Related Work}\label{relw}

We focus on state-of-the-art bivariate causal discovery methods for continuous data that are closely related to our work and highlight their identifiability results.
The first  identifiable methods are those based on the Additive Noise Model (ANM) \cite{shimizu2006linear},
where  $Y = f (X ) + N_X$  and the noise $N_X$ and cause $X$ are independent. This model
is identifiable when there does not exist an ANM in the anti-causal
direction; This holds for instance for linear functions $f$ and non-Gaussian noise $N_X$ \cite{shimizu2006linear} or for nonlinear functions and additive noise \cite{hoyer2008nonlinear} or for
post-nonlinear models \cite{zhang2009}. 
Although identifiability conditions are provided in these publications, they are often difficult to verify in practice. Furthermore, the results regarding causal direction heavily depend on the chosen independence test and the fitting algorithm  \cite{mooij2016distinguishing}.

The most related methods to
our work are  RECI from Bl\"obaum et al.  \cite{bloebaum2018}, Slope  \cite{marx2017telling}  and  Sloppy \cite{regularizedregression}, the last two from Marx and Vreeken, as these approaches base
their inference rules on the regression error. Methods based on regression
error are able to decide
between Markov equivalent DAGs under the assumption
of having a non-linear function and low noise, see  \cite{bloebaum2018}. 
 As mentioned earlier, Slope does not provide identifiability results.
Method RECI does
not employ model selection.   A third method based on  regression error is CAM from \cite{buhlmann2014cam},
which was designed to infer general causal graphs.
For the bivariate
case, CAM determines for the causal direction using regularized log-likelihood scores.

The IGCI method from  Janzing et al. \cite{janzing2012information} was also proposed for   low noise settings. The causal direction is inferred using 
the Shannon entropy of the marginals. However,  entropy estimation
for continuous data is challenging.
Method QCCD \cite{tagasovska2020distinguishing} approximates the algorithmic
Markov condition using non-parametric conditional quantile estimation. Although QCCD performs well in practice, its lacks strong
identifiability guarantees.
To the group of methods with low noise we count also  methods which  assume a homoscedastic noise.
For homoscedastic noise models,
regression
methods  are used  as model estimators in many methods, see Shimizu et al.\cite{shimizu2006linear},  Peters et al. \cite{peters2014causal}, B\"uhlmann et al. \cite{buhlmann2014cam}. Method CAM from \cite{buhlmann2014cam} also belongs among them,  using the maximum likelihood  for inferring
the cause and effect. Similarly, method  RESIT  from  Peters et al. \cite{peters2014causal}  executes the mean estimation, but an additional step of independence testing is done subsequently.

Methods which do not assume a low noise use heteroscedastic noise models, e.g. method
HECI from  Xu et al. \cite{xu2022inferring}.
HECI  partitions the domain of the
cause into multiple segments where the noise 
 is 
assumed to be homoscedastic. 
An estimator  for maximizing the Gaussian log-likelihood is presented in  method LOCI \cite{immer2023identifiability}. This method 
utilizes the log-likelihood  from the mean
and the variance parameters of the Gaussian 
distribution as a criterion (version LOCI$_{M}$)  and  the residual is recovered for the next step of
testing the independence with Hilbert-Schmidt Independence Criterion (HSIC), in version LOCI$_{H}$.
The conditions on identifiability of these LOCI versions are given,  however, similarly as for  ANMs, they are difficult to test in practice. 
There are also two neural-network  estimators of LOCI \cite{immer2023identifiability}, these without identifiability  guarantees.
Method ROCHE \cite{tran2024robust}, also without identifiability guarantees, follows the
same procedure  as
LOCI$_{H}$.  ROCHE   uses a neural network to estimate parameters of  an objective function which fits a log-likelihood of Student t-distribution on a data.
In general, the neural network estimators of causal direction have several advantages over the likelihood-based  approach, since 
they can capture  non-linear relationships between variables, and  can be more robust to noise and outliers, leading to more reliable causal direction estimates. Being aware of this advantage, we discuss the comparison of LCUBE to the methods in this group separately.



\section{Experiments}\label{exp_res}

In our experiments, we used these three groups of methods for comparison.
The regression based methods are represented by Sloppy version for splines,  since it is  most similar method to LCUBE. We do not use Slope and RECI, since they  had similar or worse results than Sloppy on all  benchmark datasets we use (see their performance in \cite{regularizedregression}).  Instead of RECI, we compare to RESIT \cite{peters2014causal}.
We used both synthetic and real-world causal discovery benchmark datasets to evaluate  LCUBE, and compared it to state-of-the-art methods.
As an evaluation metrics  we used accuracy (ACC) and  AUDRC introduced in \cite{immer2023identifiability}. 
ACC measures the fraction
of correctly inferred cause-effect relationships and the AUDRC measures the area under the decision
rate curve. 
\cite{immer2023identifiability} justifies the selection of  AUDRC over  AUROC, since it weights
correctly identified $X \to  Y$ pairs in the same way as correctly identified $Y \to X$ pairs and thus avoids an arbitrary selection of true
positives and true negatives.
(More in Appendix~\ref{audrc}.)
For  T\"ubingen pairs, for which a special weighting of each pair is given, we  used the so called forced decision from \cite{mooij2016distinguishing}. 
Our code, its  description and parameter setting are publicly available under
https://github.com/suzi216/LCUBE.


\subsection{Benchmark datasets}
We used the same benchmark datasets as in recent papers \cite{tran2024robust} and \cite{immer2023identifiability}.
The first subroup of 
synthetic datasets are from \cite{tagasovska2020distinguishing} including AN, AN-s, LS, LS-s, and MN-U.  They consist of nonlinear functions with additive noise
(AN), sigmoidal functions with additive noise (AN-s), nonlinear and
 sigmoidal location scale functions (LS and LS-s), where $Y = f(X) + g(X).N_Y$
and sigmoid functions with multiplicative uniform noise (MN-U)
with the effect Y is the form
$Y = f(X ).N_Y$.
The datasets denoted with “-s” and the MN-U dataset
use invertible sigmoid-type functions for the generative
functions which can make the identification of the causal
direction more complicated.
All these simulated datasets consist of 100 cause-effect pairs with 1 000
samples per pair.
Further we  consider
 datasets SIM,
SIM-c, SIM-ln, and SIM-G from \cite{mooij2016distinguishing}, each having 300 pairs, with  500 samples per pair.
These datasets are simulated with the functions 
$X = f_X (\varepsilon_X )$ and $Y =
f_Y (X, \varepsilon_Y)$ in the cases without a confounder (in SIM,
SIM-ln, and SIM-G) and the functions $Z = f_Z (\varepsilon_Z)$,
$X = f_X (\varepsilon_X , \varepsilon_Z )$, and $Y = f_Y (X, \varepsilon_Y , \varepsilon_Z )$ in the
cases with a confounder Z (in SIM-c). In the SIM-ln
dataset, low levels of noise are applied in the models,
and the SIM-G dataset has approximations of Gaussian distributions for the cause $X$ and approximately
Gaussian non-linear additive noise generative models.
The third group of synthetic datasets are Multi, Net from \cite{goudet2018learning}, and Cha from \cite{guyon2019}. 
The data pairs in Multi dataset are generated with pre-additive noise as
$Y = f(X + \varepsilon_Y )$, post-additive noise (similar to
conventional additive noise models), pre-multiplicative
noise $Y = f(X \varepsilon_Y )$, and post-multiplicative noise
$Y = f(X) \varepsilon_Y$. The  pairs in the Net dataset are
generated by neural networks with random distribution for $X$, such as exponential, gamma, log-normal, or
Laplace distribution.
The 
data pairs
in the Cha benchmark  are chosen from the ChaLearn
Cause-Effect Pairs Challenge \cite{guyon2019}.
All three datasets have 300 continuous variable  pairs with 500 samples per pair.
And finally,   we used the T\"ubingen dataset, consisting of 99 real-world causal pairs \cite{mooij2016distinguishing}. The sample size of Tübingen pairs varies, ranging from a few hundred to five thousand.

\subsection{Performance and comparison with the state-of-art methods}

\begin{figure*}[htbp]
    \centering
    \begin{subfigure}[t]{0.22\textwidth}
      \includegraphics[width=\linewidth]{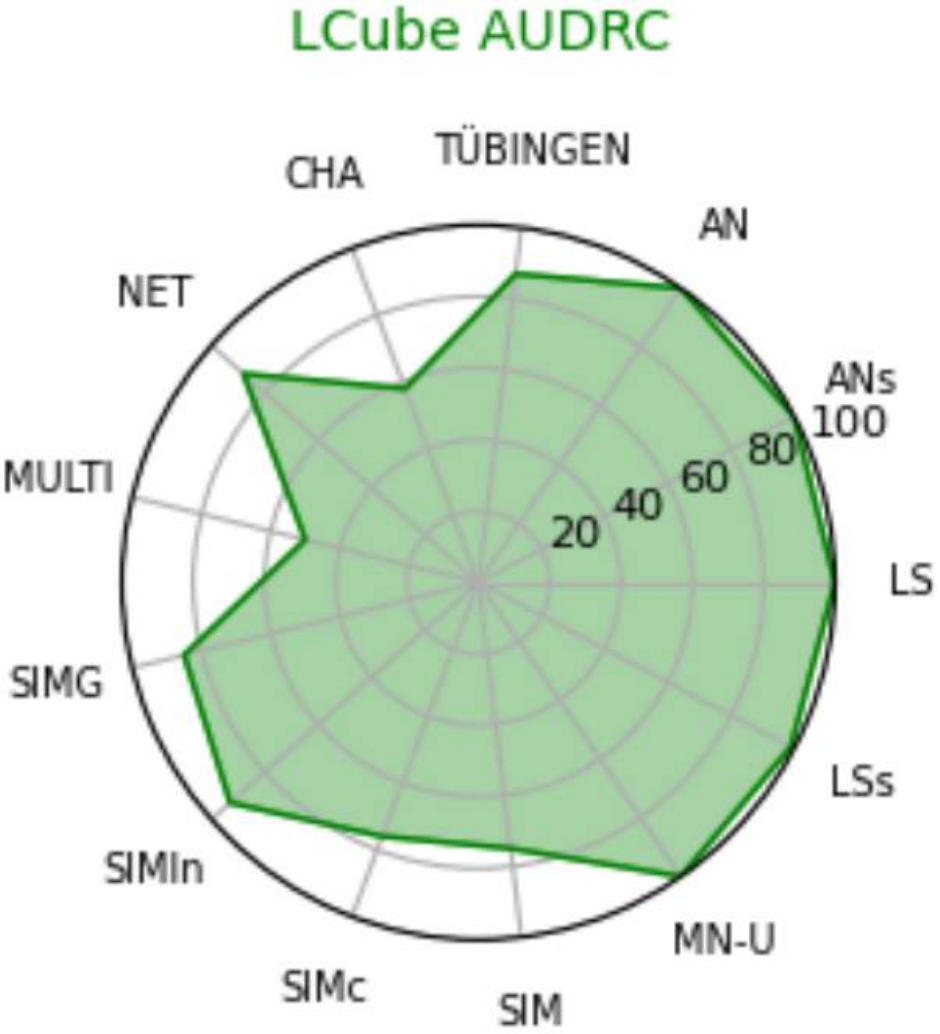}
        \caption{LCUBE}
    \end{subfigure}
    \hfill
    \begin{subfigure}[t]{0.22\textwidth}
    \includegraphics[width=\linewidth]{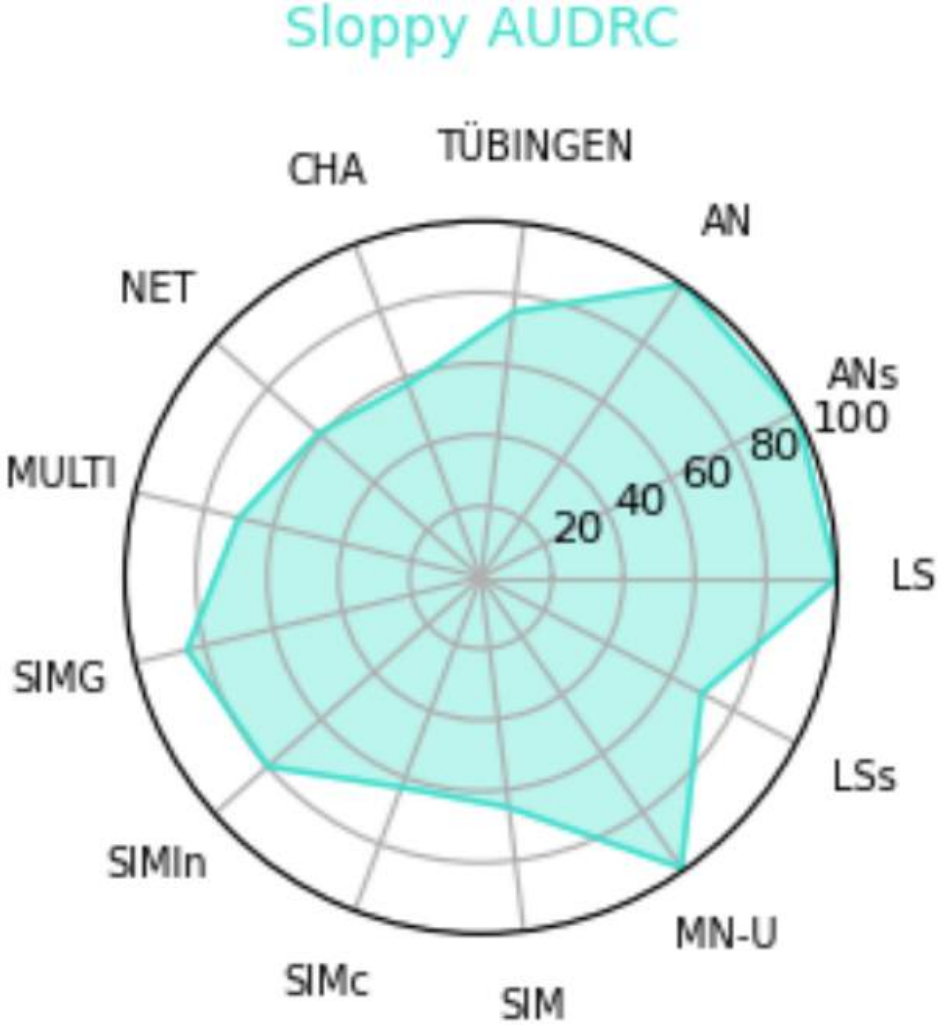}
        \caption{Sloppy}
    \end{subfigure}
    \hfill
    \begin{subfigure}[t]{0.22\textwidth}
      \includegraphics[width=\linewidth]{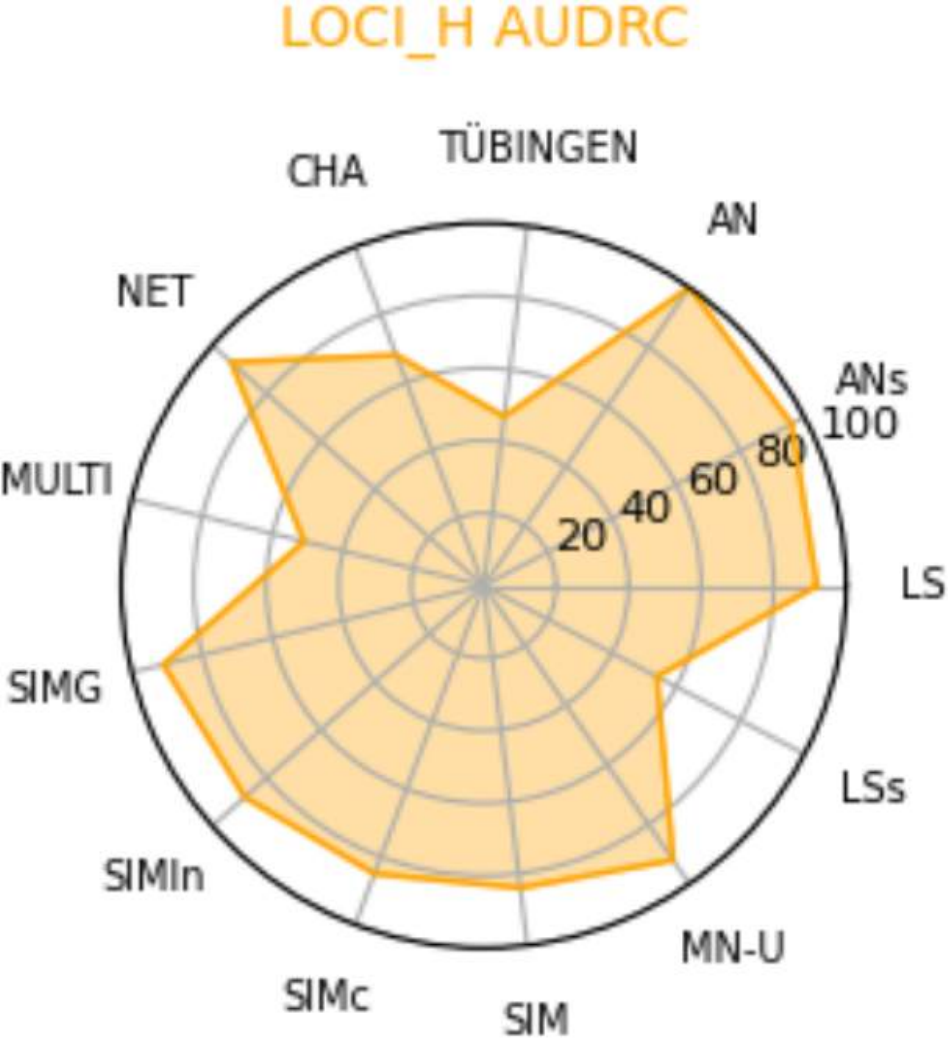}
        \caption{LOCI\_H}
    \end{subfigure}
    \centering
    \begin{subfigure}[t]{0.22\textwidth}
      \includegraphics[width=\linewidth]{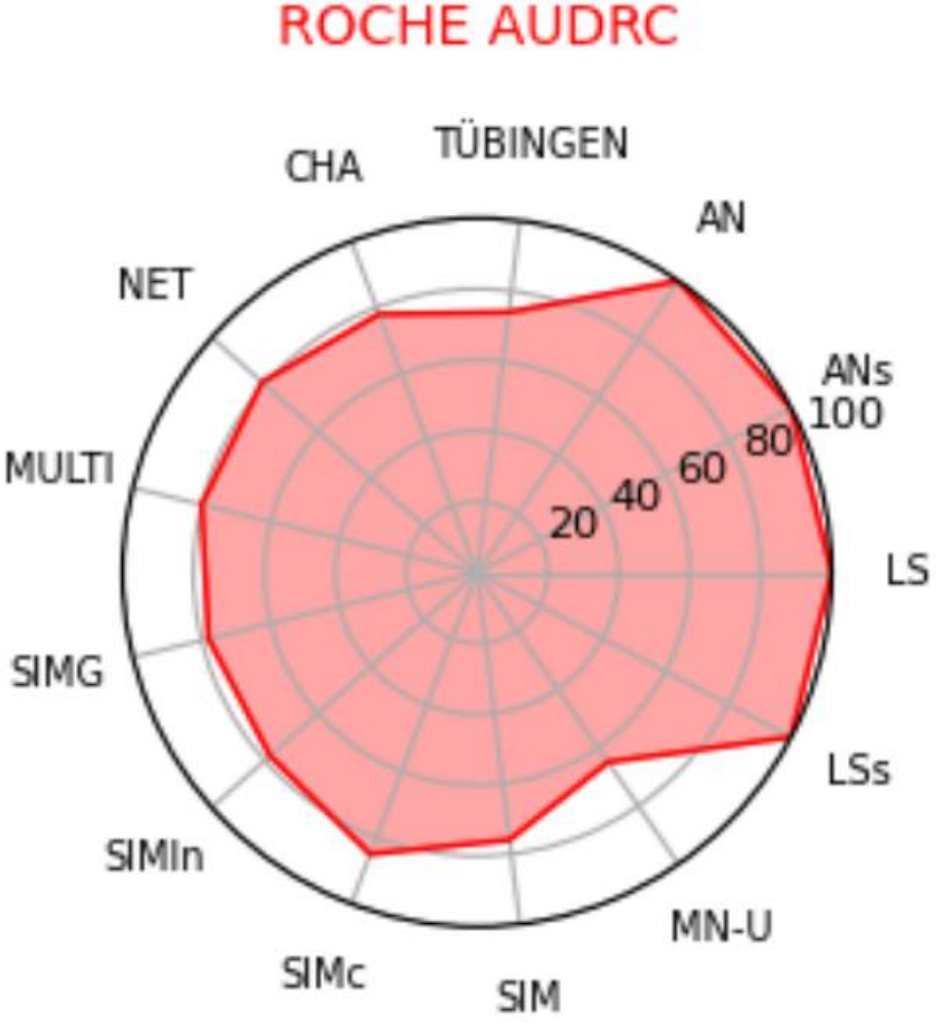}
        \caption{ROCHE}
    \end{subfigure}
      \caption{AUDRC of four methods}\label{spiderfigures}
\end{figure*}

We evaluate LCUBE with a focus on benchmark methods: 
\begin{itemize}
    \item[1.]  Methods with similar assumptions, i.e. regression with  low noise - Sloppy, CAM, QCCD, RESIT and IGCI;
    \item[2.] Methods  using regression with  heteroscedastic noise - LOCI$_M$, LOCI$_H$ and HECI;
   \item [3.] Methods  using regression with heteroscedastic noise and a neural network estimator -  NN-LOCI$_M$, NN-LOCI$_H$ and ROCHE.
\end{itemize}

The ACC and AUDRC values for  CAM, HECI,   QCCD, RESIT, IGCI,  and LOCI are adopted from  \cite{immer2023identifiability} (Table 1 and 2). 
We computed ACC and AUDRC values for ROCHE using results from their Github repository \cite{tran2024robust}. 
For Sloppy, we used the provided R code and ran it on the Cha, Net, and Multi datasets. The ACC and AUDRC values for all methods are reported in our Table~\ref{tab:final_results_Audrc_correct} and 
Table~\ref{tab:final_results_Accuracy_correct}
in Appendix~\ref{tables},
where also the performance of all methods on all 13 datasets is reported. 
The summary of the AUDRC values for LCUBE and a  representative method from each category of benchmark methods is shown in Figure~\ref{spiderfigures}.  The precisions in  AUDRC and ACC for each method in average over 10 and 13 datasets are in Table~\ref{table:audrc_acc_aver}.


\vspace{0.2cm}

\begin{table}[h!]
\centering
\begin{tabular}{|l|c|c|c|}
\hline
{ Method} &  {  AUDRC $\uparrow$}  & { Aver.AUDRC/ACC $\uparrow$
} & { Aver. AUDRC/ACC $\uparrow$} \\
\hline
\textbf{LCUBE} & \textbf{87}  & \textbf{91.5} \hspace{0.3cm} \textbf{87.5} & 85.5 \hspace{0.3cm} 81.4 \\
\hline
Sloppy & 75 & 83.7 \hspace{0.3cm} 81.1 & 78.9 \hspace{0.3cm} 73.8 \\
CAM & 70  & 79.4 \hspace{0.3cm} 78.2 & 73.8 \hspace{0.3cm} 72.5 \\
QCCD & 84  & 89.7 \hspace{0.3cm} 83.2 & 85.7 \hspace{0.3cm} 78.2 \\
RESIT & 71  & 65.7 \hspace{0.3cm} 65.0 & 68.4 \hspace{0.3cm} 64.4 \\
IGCI & 74   & 42.6 \hspace{0.3cm} 40.0 & 49.5 \hspace{0.3cm} 44.7 \\
\hline
LOCI$_M$ & 45    & 80.3 \hspace{0.3cm} 78.1 & 78.2 \hspace{0.3cm} 74.4 \\
LOCI$_H$ & 47  & 82.9 \hspace{0.3cm} 78.6 & 80.1 \hspace{0.3cm} 77.3 \\
HECI & 78 & 71.4 \hspace{0.3cm} 62.9 & 73.4 \hspace{0.3cm} 65.3 \\
\hline
{\scriptsize NN-LOCI$_M$} & 66 & 87.3 \hspace{0.3cm} 81.2 & 84.5 \hspace{0.3cm} 77.1 \\
{\scriptsize NN-LOCI$_H$} & 56   & 91.3 \hspace{0.3cm} 85.6 & 89.1 \hspace{0.3cm} 84.1 \\
ROCHE & 74 & 85.6 \hspace{0.3cm} 83.1 & 84.2 \hspace{0.3cm} 82.3 \\
\hline
\end{tabular}
\vspace{0.1cm}
\caption{
1. column: AUDRC  on real data of Tü pairs. 2. column: Average AUDRC/ACC over 10 datasets,  2. column: Average AUDRC/ACC over 13 datasets}
\label{table:audrc_acc_aver}
\end{table}

Regarding Figure~\ref{spiderfigures}, no method performs perfectly across all datasets.
However, LCUBE achieves the highest AUDRC among the four methods on the AN, ANs, LS, LSs, MN-U, and Tübingen pairs datasets.
AN, ANs, LS, LSs, MN-U have a small noise (not necessarily Gaussian).
One can see that also Sloppy, LOCI$_H$ and  ROCHE achieve a high precision
on the Gaussian AN, ANs, LS.  However, AUDRC values  of Sloppy and LOCI$_H$ drop on  LSs due to the higher complexity of this dataset. 
Sloppy's AUDRC is lower than LCUBE's on 12 datasets and it is higher only on set Multi.
LOCI$_H$ ad ROCHE show overall high AUDRC on synthetic datasets, but  significantly lower AUDRC compared to  LCUBE and Sloppy on real Tübingen data pairs.

 LCUBE achieves perfect AUDRC and accuracy on the AN, ANs, LS, LSs and MNU datasets. 
In general, the precision of the methods depends  on the precision of two main steps: the precision of an estimator for the  model and the precision of the causal score method. 
LCUBE  allows very precise estimators due to the density of the set of cubic splines. 
Interestingly, the NN  method
ROCHE demonstrates  high precision across all synthetic sets, except for  MN-U, where it has ACC of 62  and AUDRC  of 65. In contrast,  LCUBE has  perfect precision (100) on MN-U.  While all NN methods  perform well on synthetic datasets, they lack   identifiability guarantees. They also  perform poorly on real data of Tübingen pairs, with 
 AUDRC  of 56 for NN-LOCI$_H$ and  66  (NN-LOCI$_H$), and  74 (ROCHE), while AUDRC of LCUBE achieves  score 87. 
 To the best of our knowledge, this is the highest score achieved by any state-of-the-art method. As stated before, LCUBE is a scoring method with identifiability guarantees.

 LCUBE has only one hyper-parameter: the upper bound  for number of knots $m_{max}$, which is also used by Sloppy. In contrast,   Roche, NN-LOCI$_H$ and NN-LOCI$_M$ require multiple   hyperparameters for their  NN architectures and  Adam optimizer. Methods that use HSIC score require additional hyperparameters, including the kernel type, bandwidth and regularization parameter.
To accelerate the execution of the computational models, we parallelized both the  code and  the data. We used the  NVIDIA  RTX 4090 GPU and 64 GB of  RAM. 
Regarding the real running time on our hardware, LCUBE is significantly faster than both Sloppy and ROCHE. 
For example, on the Tübingen pairs dataset, LCUBE takes approximately 3  minutes, while Sloppy takes 20 minutes  and ROCHE around 6  hours.
On the Cha, Net or Multi datasets, 
LCUBE takes 5 minutes  and 2 minutes on the other  datasets. In comparison, Sloppy takes 15 min on Cha, Net, Multi, and 10 min on  AN, MN-U, SIM. 
Roche requires around 8 hours on  Cha, Net, and Multi and 5 hours on AN, MN-U, and SIM.

\section{Conclusions}\label{conclusions}
In this work, we proposed a bivariate causal score based on the the Minimum Description Length (MDL)   principle, using functions  with the density property  and  proved the identifiability of these  scores under specific conditions. 
These conditions can be easily tested. Gaussianity of the noise in the causal model equations is not assumed, only that the noise is low.
 We also introduced LCUBE as
an instantiation of the MDL-based causal score utilizing well-studied cubic
regression splines. 
LCUBE is  identifiable,  
interpretable, simple and very fast.
LCUBE has only one hyperparameter.
 It achieves superior precision in terms of AUDRC on the Tübingen
cause-effect pairs dataset  compared to the
state-of-the-art methods. 
It also shows superior average precision across common 10 benchmark 
datasets and above average
precision on 13 datasets. 
In future work, we aim to explore  causal scores via other  dense functional classes.

\vspace{0.2cm}

\noindent \textbf{Acknowledgements}\\
We thank Dr. Negar Safinia naini  and Dr. Amir Rahnama for valuable discussions  in the early phase of the paper and Prof. Jilles Vreeken for his comments  in the final phase.


\appendix

\subsection{Proof of Lemma 1:}\label{proofL1}
\noindent Under Assumptions 1-3 from Blöbaum et al. in \cite{bloebaum2018}, it holds 

\begin{equation}\label{bloebaum_ratio}
\lim_{\alpha \to 0} \frac{\mathbb{E} [ (\tilde{Y}_\alpha - \varphi(X))^2 ]}
{\mathbb{E} [ (X - \psi (\tilde{Y}_\alpha))^2 ]} \le 1 
\end{equation}

with equality only if the $\phi$ is linear. (Note: The fraction on the left side of Eq.~(\ref{bloebaum_ratio}) is in Theorem 1 in \cite{bloebaum2018} is formulated as $\lim_{\alpha \to 0} \frac{\mathbb{E} [ ( X- \psi(\tilde{Y}_\alpha )^2 ]}
{\mathbb{E} [ (\tilde{Y}_\alpha - \phi (X)^2 ]} \ge 1$ and thus this inequality is equivalent to Eq.~ (\ref{bloebaum_ratio}).)
Since Eq.~(\ref{bloebaum_ratio}) holds for any $\alpha \to 0$, it holds also  for  model given by Eq.~(\ref{ANM_alpha3}) and thus we can write

\begin{equation}\label{bloebaum_ratio_n}
\lim_{n \to \infty} \frac{\mathbb{E} [ (\tilde{Y}_n - \varphi(X))^2 ]}
{\mathbb{E} [ (X - \psi (\tilde{Y}_n))^2 ]} \le 1. 
\end{equation}


\noindent Similarly as in proof of Theorem 1 in \cite{regularizedregression},  
as $S$ is an IRSF, we can write it as 
$S(a, b) := \gamma (a) + \lambda(b)$
where $\gamma$ is a strictly  increasing function.  The statement does not change by applying $\gamma$ to the numerator and denominator in Eq.~(\ref{bloebaum_ratio_n}). 

\noindent Based on Assumption 4, i.e.  
$\|\beta_{\varphi} \|_0 \leq \|\beta_{\psi} \|_0$ holds and thus 
\begin{equation}
\frac{\gamma ( \mathbb{E} [ (\tilde{Y}_n - \varphi(X))^2 ]) + \|\beta_{\varphi} \|_0}
{\gamma ( \mathbb{E} [ (X - \psi (\tilde{Y}_n))^2 ]) + \|\beta_{\psi} \|_0}
\leq 1, \nonumber
\end{equation}
with equality if and only if $\|\beta_{\varphi} \|_0 = \|\beta_{\psi} \|_0$.
As  $\lambda$  is strictly  increasing, applying it to  $\|\beta_{\varphi} \|_0 \text{ and } \|\beta_{\psi} \|_0$ will not change this statement.

 \begin{table*}[t]
\centering
\begin{tabular}{c|c|ccccc|cccc|ccc}
\hline
& T\"u & AN & ANs & LS 
&  LSs  &  MNU  &  SIM  &  SIMc  & SIMln  &  SIMG  &  Multi  &  Net &  Cha\\ \hline
 LCUBE       & 87   &  100    &  100    &  100     &  99  & 100    & 75   &  76   & 93  & 85   & 50 &  88   & 58   \\ 
 \hline
  Sloppy  &  75    & 100  & 100  & 100 &  70 & 99 & 65 & 63 & 80 & 85 & 70 & 61 & 58\\
 CAM &  70    &  100  & 100  & 100 & 36  & 76 & 68 & 69 & 87 & 88 & 85 & 40  &  41\\ 
QCCD &   84   &  100  &  91 & 100 & 100 & 100 & 71 & 83 & 92 & 76  & 63 & 94 &  61\\ 
RESIT &  71    & 100  & 100  & 69 & 4 & 0 & 75  & 84  & 83 & 71 & 68 & 81 & 83 \\ 
IGCI &  74    & 18   & 35  & 60  & 49 &  1 &  34 & 41 & 51 & 63  & 99  & 61  & 58 \\ 
\hline
LOCI$_M$ &  45    &  98  &  95  & 88 & 86 & 90 &  68 &  56 & 90 & 87 &  75 &  84 &  55\\ 
LOCI$_H$ &   47   & 100   & 96  & 92 & 54 &  92 &  84 & 85 &  88 & 91 & 51 & 93 & 68 \\ 
HECI &  78    &  100  & 63  & 99 & 72 & 20 & 59 &  64 & 86 & 73 & 99 & 84 & 57\\ 
\hline
NN-LOCI$_{M}$ &   66   & 100   & 100   & 100 & 100 & 100 & 60 & 63 & 95 & 89 & 93 & 86 & 47\\ 
NN-LOCI$_{H}$ &  56    & 100   & 100  & 99 & 97  & 100 & 89 & 93 & 86 & 93 & 77 & 97 & 71 \\ 
ROCHE & 74 & 100 &  100 & 100 & 100 & 65 & 76 & 85  & 78 & 78 &  80 & 81 &  78\\
\hline
\end{tabular}
\vspace{0.1cm}
\caption{ 
AUDRC in $\%$  for concrete datasets and methods. CAM to NN-LOCI$_{H}$ are from \cite{immer2023identifiability}.}
\label{tab:final_results_Audrc_correct}
\end{table*}
\begin{table*}[t]
\centering
\begin{tabular}{c|c|ccccc|cccc|ccc}
\hline
& T\"u & AN & ANs & LS 
&  LSs  &  MNU  &  SIM  &  SIMc  & SIMln  &  SIMG  &  Multi  &  Net &  Cha\\ \hline
 LCUBE       & 72   &  100    &  100    &  100     &  98  & 100    & 67   &  68   & 90  & 80   & 45 &  81   & 57   \\ 
 \hline
  Sloppy  &  76    & 100  & 100  & 100 &  56 & 96 & 64 & 62 & 77 & 81   & 50 & 46 & 52  \\
 CAM &  58    &  100  &  100 & 100 & 53 & 86 & 57 & 60 & 87 & 81  & 35 &  78 & 47 \\ 
QCCD &   77   &  100  &  82 & 100 & 96 & 99 & 62 & 72  & 80 &  64 & 51 &  80 & 54 \\ 
RESIT &  57    & 100  & 100  &  61 & 6  & 2 &  77 & 82 & 87 & 78 & 37 & 78 & 72 \\ 
IGCI &  68    & 20   &  35 & 46 & 34 & 11 & 37 & 45  & 51 & 53 & 92 &  35 & 55 \\ 
\hline
LOCI$_M$ &  52    &  99  & 98   & 94 & 94 &  93 & 52  &  48 & 77 & 74 & 66  & 75  & 46 \\ 
LOCI$_H$ &   56   &  99  & 98  & 85 & 53 & 90 & 75 & 76 & 73 & 81 & 66 & 84 & 70 \\ 
HECI &  71    &   98 &  55 & 92 & 55 & 33 & 49 & 55 & 65 &  56 & 91 & 72 & 57\\ 
\hline
NN-LOCI$_{M}$ &   57   &  100  & 100   & 100 & 100 & 100 & 48 & 50 & 79 &  78 & 72 & 76 &  43\\ 
NN-LOCI$_{H}$ &  60    & 100   & 100  & 95 &  89 & 100 & 79 & 83 & 72 & 78 & 78 & 87 & 72 \\ 
ROCHE & 70 & 100 & 100  & 100 & 100 & 62 & 71 &  80 & 74 & 74 & 78 & 81 & 80 \\
\hline
\end{tabular}
\vspace{0.1cm}
\caption{
Accuracy in $\%$ for concrete datasets and methods. CAM to NN-LOCI$_{H}$ are from \cite{immer2023identifiability}.}
\label{tab:final_results_Accuracy_correct}
\end{table*}

\subsection{Code length $L(\btheta)$  for a cubic regression spline}\label{derivation_Ltheta}

For the moment  we select  a fixed number of (internal) knots  $m = |\bk|$ where 
$\bk =\{k_1, \dots, k_{m}\} \subset X$.

\subsubsection{Code $L(m)$} 
$L(m)$ can be approximated by $\log m$ when $m$ is  reasonably large, see  paper \cite{rissanen1998stochastic}, Section 2.2.4. Thus based on this,
$L(m) \approx \log_2 m.$
\vspace{0.1cm}

\subsubsection{Code $L(\bk|m)$}\label{coding_betas}

First we will compute the code $L(\bk|m)$.
Since $\bk =\{k_1, \dots, k_{m}\} \subset X$, 
the sequence of indices $\bk$ can be specified by the indices of those $x_i$'s  where a knot is placed. 
This set of sorted indices can be compactly specified by their
successive differences.
Define $k_0 = \min(x_i)$
and $k_{m+1} = \max (x_i)$
and similarly as in \cite{lee2000regression} let 


 \begin{equation}\label{u_j2}
 u_j :=  \mbox{the  number of } x_i's \mbox{ so that }  k_{j-1} \le x_i \le k_j \nonumber
 \end{equation}


for $j=1, \dots, m$. 
Each $u_j$ is the $j-th$ successive (i.e. next) index difference.  If we know the whole sequence $(u_1, \dots, u_{m})$,  we also know the whole knot sequence $\bk$. 
 If the knots $\bk$ satisfy the Schoenberg-Whitney conditions, is each $u_j$  a positive  integer and the sequence $\bk$ can be coded by 

\begin{equation}
L(\bk|m) = L(u_1, u_2, \dots, u_{m}|m )= \sum_{j=1}^{m} L(u_j) \nonumber
\end{equation}
 
where $L(u_j) =  \log u_j$. 
Since $u_j$ are integer values,  holds

\begin{equation}
L(\bk|m) = 
L(u_1, \dots, u_1|m)
\approx
\sum_{j=1}^{m} \log u_j.\nonumber
\end{equation}


\subsubsection{Code for $L(\bb, \bbeta| m, \bk)$}

Now we compute $L(\bb,\bbeta| m, \bk)$.
Given $m$, $\bk$, the values    $\bb, \bbeta$ can be computed from 
Eq.~(\ref{Eq:beta_estimate}) and  the result is 
the conditional maximum likelihood estimates of $\bb$ and $\bbeta$
(or the least square estimates if the assumption of normal errors is not given). 
  Based on  \cite{rissanen1998stochastic},  pp. 55-56,  if a conditional max. lik. estimate is estimated from $n$ points, then it can be efficiently encoded with $\frac{1}{2} \log n$.
      Since $r = 3$, one can see that  each $b_j's$ and $\beta_j's$ is estimated  from all $n$ x-values.
Hence

\begin{equation}
L(b_0) = \dots = L(b_{3}) = L(\beta_1 )  = \dots L(\beta_{m} ) =  \frac{1}{2} \log n \nonumber
\end{equation}
Since a cubic  spline  is defined by $4$ coefficients then 
$
L(\bb,\bbeta| m, \bk) =   \frac{m+ 4}{2} \log n.
$
To summarize,
$$
L(\btheta) = \log m   
+ \sum_{j=1}^{m} \log u_j + \frac{m+ 4}{2} \log n.$$


\subsection{Complexity of LCUBE}\label{complexityLCUBE}
The placement of $m$ knots  has a comp. cost of $O(1)$. Constructing the spline basis  in Eq.~(\ref{Eq:regression_spline}) requires $O(m)$ operations. Solving the least squares for $n$ samples and $m$ basis functions, which involves solving an $m \times m$ linear system   with $n \times m$ design matrix, results to  total  costs $O(nm + m^3)$.


\subsection{Evaluation measures}\label{evaluation}
We use  evaluation metrics accuracy and  AUDRC from \cite{immer2023identifiability}. 
For real data of T\"ubingen pairs,  we  use the so called forced decision as defined in \cite{mooij2016distinguishing}. 

\subsubsection{AUDRC}\label{audrc}
The area under the decision
rate curve (AUDRC).
\cite{immer2023identifiability} justifies the selection of  AUDRC over  AUROC, since it weights
correctly identified $X \to  Y$ pairs in the same way as correctly identified $Y \to X$ pairs and thus avoids an arbitrary selection of true
positives and true negatives, which can lead to non-interpretable
results on unbalanced datasets, see   \cite{marx2017telling},  Table 1.
The accuracy  (ACC) measures the fraction
of correctly inferred cause-effect relationships and the AUDRC measures how well the decision certainty indicates
accuracy. The certainty is, for example, indicated by the
likelihood or $p$-value difference in both directions. Thus, a
high AUDRC indicates that an estimator tends to be correct
when it is certain and only incorrect when it is uncertain.
Given the ground truth direction 
$t(.)$ and
a causal estimator for the direction $f(.)$ on $Q$ pairs with
ordering $\pi(.)$ according to the estimator’s certainty, 
$$
  AUDRC = \frac{1}{Q}\sum_{q=1}^Q \frac{1}{q} \sum_{i=1}^q\mathbf{1}_{f(\pi(i))=t(\pi(i))},
$$
the indicator function \( \mathbf{1}_A(x) \) is defined as 1 if $x \in A$ otherwise 0.
I.e.,  we average the accuracy when iteratively adding
the pair the estimator is the most certain about. 
Higher AUDRC means
that a method predicts the correct causal directions if the confidence of the scores is high.


\subsubsection{Forced decision: Evaluation of accuracy ACC}
 Given  $(X,Y)$, the method has to decide either
$X \to Y$ or $Y \to X$;
We evaluate the accuracy of these decisions as
$$ accuracy =  \frac{\sum_{q=1}^Q w_q \delta_{\hat{d}_q, d_q}}{\sum_{q=1}^Q w_q},$$
where
$d_q$ is the true causal direction for the q’th T\"ubingen pair (either $\leftarrow$  or $\rightarrow$), and $\hat{d}_q$ is the
estimated direction (i.e.  $\leftarrow$  or $\rightarrow$ or ? = undecided) and $w_q$ is the weight of the pair, see \cite{mooij2016distinguishing}.

\subsection{Performance of all methods}\label{tables}
\noindent AUDCRC and accuracy in $\%$ for each method and dataset can be found in 
Table~\ref{tab:final_results_Audrc_correct} and Table~\ref{tab:final_results_Accuracy_correct}.


\end{document}